\documentclass[conference]{IEEEtran}
\IEEEoverridecommandlockouts

\usepackage{cite}
\usepackage{amsmath,amssymb,amsfonts}
\usepackage{algorithmic}
\usepackage{graphicx}
\usepackage{textcomp}
\usepackage{xcolor}
\def\BibTeX{{\rm B\kern-.05em{\sc i\kern-.025em b}\kern-.08em
T\kern-.1667em\lower.7ex\hbox{E}\kern-.125emX}}
\begin{document}

\title{The Energy-Efficient Hierarchical Neural Network with Fast FPGA-Based Incremental Learning\\
\thanks{This material is based upon work supported by the National Science Foundation under Grant No. 2234227.}
}

\author{\IEEEauthorblockN{Mohammad Saleh Vahdatpour}
\IEEEauthorblockA{\textit{Department of Computer Science} \\
\textit{Georgia State University}\\
Atlanta, GA 30302-5060, USA \\
msvahdatpour@gmail.com}
\and
\IEEEauthorblockN{Huaiyuan Chu}
\IEEEauthorblockA{\textit{Department of Computer Science} \\
\textit{Georgia State University}\\
Atlanta, GA 30302-5060, USA\\
chuhy4@gmail.com}
\and
\IEEEauthorblockN{Yanqing Zhang}
\IEEEauthorblockA{\textit{Department of Computer Science} \\
\textit{Georgia State University}\\
Atlanta, GA 30302-5060, USA\\
yzhang@gsu.edu}
}

\maketitle

\begin{abstract}

The rising computational and energy demands of deep learning, particularly in large-scale architectures such as foundation models and large language models (LLMs), pose significant challenges to sustainability. Traditional gradient-based training methods are inefficient, requiring numerous iterative updates and high power consumption. To address these limitations, we propose a hybrid framework that combines hierarchical decomposition with FPGA-based direct equation solving and incremental learning. Our method divides the neural network into two functional tiers: lower layers are optimized via single-step equation solving on FPGAs for efficient and parallelizable feature extraction, while higher layers employ adaptive incremental learning to support continual updates without full retraining. Building upon this foundation, we introduce the Compound LLM framework, which explicitly deploys LLM modules across both hierarchy levels. The lower-level LLM handles reusable representation learning with minimal energy overhead, while the upper-level LLM performs adaptive decision-making through energy-aware updates. This integrated design enhances scalability, reduces redundant computation, and aligns with the principles of sustainable AI. Theoretical analysis and architectural insights demonstrate that our method reduces computational costs significantly while preserving high model performance, making it well-suited for edge deployment and real-time adaptation in energy-constrained environments.
\end{abstract}

\begin{IEEEkeywords}
sustainable artificial intelligence, hierarchical decomposition, software-hardware co-designed learning, FPGA, incremental learning, large language models, foundation models.

\end{IEEEkeywords}

\section{Introduction}
Artificial Intelligence (AI) has revolutionized various fields, from healthcare and autonomous systems to finance and scientific computing, driving major advances in automation, predictive analytics, and decision-making. However, the increasing computational demand of AI models—especially deep learning architectures—poses significant sustainability challenges. The growing reliance on large-scale neural networks has led to substantial energy consumption and environmental costs, as training and deploying these models requiring vast computational resources, often relying on carbon-intensive infrastructures. Large language models (LLMs), in particular, have exacerbated these challenges, as their rapid growth demands increasingly complex hardware configurations, resulting in high memory and power consumption \cite{2501.11006v1, 2111.05193v2}. Studies have shown that training a single transformer-based model can emit as much carbon as five cars over their lifetime, highlighting the urgent need for sustainable AI solutions that minimize energy waste while maintaining high model performance \cite{MLSys-2022-sustainable-ai-environmental-implications-challenges-and-opportunities-Paper}.

A primary source of inefficiency in current AI systems is the reliance on traditional gradient-based optimization methods, which involve iterative weight updates and imposes substantial computational and memory overhead. These training processes require extensive memory and processing resources, increasing both power consumption and hardware costs. AI hardware, particularly GPUs and TPUs, typically operates at high power levels, exacerbating environmental concerns. The issue is particularly severe for LLMs, as their sheer scale leads to memory bottlenecks and inefficient parameter updates. Recent research has explored techniques such as FlashAttention, quantization, and sparsity-aware training to mitigate these challenges, but fundamental limitations persist in large-scale AI training \cite{LLM-in-a-flash, 2205.09646v1}. Moreover, the data movement between memory and processing units in these architectures leads to additional energy losses, making the entire pipeline inefficient. Addressing these bottlenecks calls for a paradigm shift towards energy-efficient AI, integrating innovative training methodologies with hardware-aware optimization techniques \cite{Green_Edge_AI_A_Contemporary_Survey}.

One promising avenue for achieving sustainability in AI is software-hardware co-design, where algorithmic enhancements are co-developed with optimized hardware architectures to improve energy efficiency and computational throughput. Unlike conventional AI pipelines that treat software and hardware separately, co-design approaches ensure tight integration between model optimizations and hardware constraints. LLMs stand to benefit significantly from such co-design principles, as recent work has shown that fine-tuning large-scale models with hardware-aware optimizations can substantially reduce energy usage \cite{llm_survey_2303.18223}. Recent research in Green Granular Neural Networks (GGNNs) has demonstrated that direct equation solving on FPGA—rather than traditional gradient descent—can significantly reduce computational overhead, lowering both power consumption and training time \cite{chu2023green, chu2023green2}. FPGA-based computation allows for parallel processing and customized logic operations, making it an attractive alternative to traditional AI accelerators \cite{HASCO}. However, one of the key challenges in FPGA-driven AI acceleration is scalability. While direct equation solving is efficient for small to mid-sized networks, its computational complexity grows exponentially with model size, limiting its applicability to large-scale architectures such as CNNs, transformers, and deep reinforcement learning models \cite{On_the_Viability_of_Using_LLMs_for_SW_HW_Co-Design_An_Example_in_Designing_CiM_DNN_Accelerators}.

To address these challenges, we propose a hybrid learning framework that integrating hierarchical decomposition, FPGA-based direct equation solving, and incremental learning to balance efficiency and adaptability. This is particularly crucial for LLMs and foundation models, where training costs are prohibitive and incremental learning methods can significantly reduce computational overhead \cite{2408.15518v2}. In this approach, lower network layers, responsible for extracting fundamental features, leverage FPGA-based computation for rapid and energy-efficient training. Higher layers, which capture more complex and abstract representations, employ incremental and adaptive learning strategies. This hybrid framework allows models to dynamically update without full retraining, thus reducing computational costs and ensuring sustainability. Additionally, the approach mitigates numerical instability by selectively applying direct equation solving where it is computationally feasible, while leveraging adaptive learning to preserve flexibility and scalability. By combining these strategies, our method seeks to enhance the efficiency of AI models while minimizing their carbon footprint and hardware requirements. Furthermore, we introduce the Compound LLM structure, inspired by agent-based AI systems and modular LLM designs, which explicitly deploys LLM modules at both hierarchical levels. This architecture enhances the clarity, modularity, and practical deployment of our sustainable framework for large-scale language models.
In this paper, we extend sustainable AI research by applying hierarchical decomposition principles to foundation models and LLMs. Given the growing dominance of LLMs across various applications, optimizing their training and inference phases with energy-efficient techniques is critical. Our work builds on recent advancements in LLM memory optimization, quantization, and adaptive fine-tuning, ensuring that large-scale models can be trained and deployed sustainably without sacrificing performance. We first conduct a literature review on recent advancements in sustainable AI, covering techniques such as low-power hardware acceleration, energy-efficient training paradigms, and hardware-algorithm co-optimization. We then present our proposed methodology, outlining its theoretical foundations and implementation details. Finally, we analyze the theoretical implications of our approach, evaluate its feasibility for real-world deployment, and discuss its potential contributions to sustainable AI research and energy-efficient computing systems.

The primary contributions of this work are summarized as follows:
\begin{itemize}
\item We propose a hybrid framework that integrates hierarchical decomposition with FPGA-based direct equation solving and incremental learning to enhance energy-efficient AI model training.
\item We introduce the Compound LLM architecture, which explicitly applies modular energy-efficient LLMs across two functional tiers for scalable deployment.
\item We address the scalability challenges of FPGA-based methods for large models and propose hybrid optimization strategies for future extensions.
\item We provide theoretical analysis and design insights highlighting how our method improves adaptability, reduces computational redundancy, and advances sustainable AI research.
\end{itemize}

\section{Sustainable AI}
The rapid growth of artificial intelligence (AI) has introduced significant energy consumption challenges, especially in deep learning applications where large-scale models require vast computational resources. The carbon footprint of training deep networks, such as transformer-based architectures, has been estimated to reach the equivalent of five cars’ lifetime emissions, highlighting the need for more sustainable AI methods \cite{MLSys-2022-sustainable-ai-environmental-implications-challenges-and-opportunities-Paper}. Such environmental impact is largely driven by traditional gradient-based optimization methods, which demand continuous weight updates and high memory usage, leading to excessive power consumption \cite{Green_Edge_AI_A_Contemporary_Survey}. Addressing these challenges calls for a paradigm shift toward energy-efficient AI, combining innovations in algorithm design with hardware-aware acceleration strategies.

Sustainable AI focuses on optimizing AI systems at multiple levels, including data efficiency, model compression, and hardware-aware execution. Green AI frameworks incorporate best practices across these three domains—data-centric, model-centric, and system-centric optimization—to minimize computational waste while maintaining performance \cite{2406.18142v1}. This multi-level approach enables AI models to balance accuracy with energy efficiency, lowering the environmental footprint of large-scale training and inference. With the rise of foundation models and LLMs, sustainable AI methods must now address not only traditional deep learning models but also the immense computational costs associated with pretraining and fine-tuning large-scale transformers \cite{2205.09646v1, 2501.11006v1}.

One promising direction in sustainable AI research is model compression, aiming to reduce computational load while preserving model accuracy. Techniques such as quantization, pruning, and knowledge distillation have been extensively explored to lower memory and processing requirements \cite{2406.18142v1}. Additionally, edge AI solutions allow inference to be performed directly on low-power devices, reducing reliance on centralized cloud infrastructure and minimizing energy consumption \cite{analytics-03-00008}. However, while these methods offer partial solutions, they do not fully resolve scalability challenges posed by increasingly complex AI models. In the case of LLMs, model compression techniques such as FlashAttention and mixture-of-experts architectures have been explored to make large models more computationally efficient without significantly compromising performance \cite{LLM-in-a-flash}.

Another critical area of sustainable AI research involves optimizing hardware architectures for AI workloads. Traditional GPUs and TPUs are designed for high-throughput computations but often fail to maximize energy efficiency. Recent advances in FPGA-based AI acceleration demonstrate significant improvements in power efficiency, as these reconfigurable hardware platforms support parallel processing tailored to specific AI tasks \cite{HASCO}. Neuromorphic computing and processing-in-memory (PIM) architectures also present promising alternatives, aiming to reduce energy waste by minimizing data movement and leveraging memory-efficient computations \cite{2310.09385v2}. For foundation models, researchers are now exploring FPGA-based acceleration for transformer architectures, enabling efficient training and inference without the need for high-power GPUs \cite{2111.05193v2}.

In addition to hardware innovations, algorithmic strategies such as adaptive training schedules and federated learning frameworks are being explored to further reduce energy consumption. These approaches dynamically allocate computational resources based on workload demands, ensuring that training and inference operate efficiently under varying conditions \cite{Green_Edge_AI_A_Contemporary_Survey}. Additionally, carbon-aware AI scheduling, where model training is aligned with periods of low-carbon energy availability, has been proposed as a promising strategy to mitigate the environmental impact of large-scale training operations \cite{MLSys-2022-sustainable-ai-environmental-implications-challenges-and-opportunities-Paper}.

Despite recent progress, the challenge of maintaining AI sustainability at scale remains an open challenge. While compression and efficient hardware improve computational efficiency, they often introduce trade-offs in model accuracy and adaptability. Similarly, FPGA and neuromorphic approaches require specialized hardware implementations, limiting their broad adoption. A more integrated strategy, combining efficient learning strategies with hardware-aware design, is needed to ensure long-term sustainability. For LLMs, sustainable training requires balancing energy-efficient architectures with scalable optimization techniques that preserve accuracy and robustness \cite{llm_survey_2303.18223}.

In this work, we aim to contribute to this area by leveraging software-hardware co-design and hierarchical decomposition to enhance energy-efficient AI. By integrating direct equation solving with adaptive learning techniques, our proposed method seeks to balance computational efficiency and model adaptability. This framework is particularly relevant for foundation models and LLMs, where efficient training methods are essential for reducing carbon footprints while maintaining high performance in real-world applications. The following sections explore how these approaches can be combined to achieve scalable, low-power AI and LLMs while maintaining high performance.

\section{Optimizing AI and LLMs Through Software-Hardware Co-Design}

AI models have traditionally been developed using a software-first approach, where algorithms are designed independently of the underlying hardware. This separation often leads to inefficient resource utilization, as software optimizations may fail to fully exploit the capabilities of the hardware. Software-hardware co-design addresses this challenge by jointly optimizing AI algorithms and hardware architectures to maximize efficiency, reduce power consumption, and improve computational performance \cite{HASCO}. With the emergence of foundation models and LLMs, software-hardware co-design has become increasingly crucial. Optimizing these models requires integrating hardware-aware strategies that improve inference efficiency and reduce the energy cost of training and fine-tuning large-scale architectures \cite{2205.09646v1, llm_survey_2303.18223}.

One of the most promising platforms for sustainable AI acceleration is Field-Programmable Gate Arrays (FPGA). Unlike GPUs and TPUs, which utilize fixed architectures, FPGAs allow for custom hardware configurations tailored to specific AI workloads. This flexibility enables energy-efficient implementations of deep learning models by optimizing data flow, reducing redundant computations, and leveraging parallel processing \cite{2410.07265v1}. For instance, direct equation solving on FPGA, instead of traditional backpropagation, has demonstrated significant reductions in power consumption and training time while maintaining model accuracy. However, despite these advantages, FPGAs face challenges in scalability and accessibility due to the complexity of hardware design and the need for specialized programming knowledge \cite{Fire-Flyer_AI-HPC_A_Cost-Effective_Software-Hardware_Co-Design_for_Deep_Learning}. Recent research has explored FPGA-based acceleration for transformers and LLMs, demonstrating that model-specific FPGA optimizations can significantly enhance both training and inference performance \cite{2111.05193v2}.

Another promising direction in AI hardware efficiency is neuromorphic computing, which emulates the brain’s parallel processing capabilities to perform computations with exceptionally low energy overhead. Unlike traditional von Neumann architectures, neuromorphic processors utilize spiking neural networks (SNNs) to process information asynchronously, reducing power consumption while maintaining real-time inference capabilities. Although current neuromorphic chips show promise, their broader adoption is limited by minimal software support and the need for novel learning paradigms tailored to SNNs \cite{2412.08490v1}.

While these hardware advancements present promising opportunities, effective software integration remains crucial for maximizing their impact. Algorithm-hardware co-optimization ensures that AI models are structured in a way that exploits hardware-specific advantages, such as sparsity-aware computation, mixed-precision training, and memory-efficient neural network architectures. For instance, quantization techniques can significantly reduce the computational requirements of neural networks while preserving accuracy, making them more compatible with low-power accelerators \cite{3478684.3479261}. Similarly, techniques like model pruning remove redundant connections in neural networks, enabling compact models that require fewer resources for inference \cite{2403.14635v1}. LLMs, in particular, benefit from techniques such as FlashAttention, mixed-precision training, and model sparsification, which help reduce their computational burden while preserving accuracy \cite{LLM-in-a-flash}.

Despite these advancements, scalability remains a major challenge. Many software-hardware co-design techniques are effective for small to mid-sized models but become inefficient when extended to large-scale architectures such as transformers and deep reinforcement learning models. For LLMs, memory constraints and high computational costs necessitate novel approaches such as hardware-efficient attention mechanisms and FPGA-adapted training pipelines to improve processing efficiency while lowering energy consumption \cite{2501.11006v1}. To address this, we propose a hierarchical decomposition strategy that combines direct equation solving for lower network layers with adaptive incremental learning for higher layers. This approach allows efficient hardware execution while maintaining adaptability, leading to a more sustainable AI framework.

The next section examines hierarchical decomposition for energy-efficient AI and LLMs, detailing its role in enhancing the scalability, adaptability, and sustainability of deep learning systems and large-scale language models.

\section{Hierarchical Decomposition for Energy-Efficient AI and LLMs}

The computational inefficiencies of deep learning models, as discussed in the previous sections, underscore the growing need for sustainable AI solutions. While software-hardware co-design optimizes computational efficiency at the hardware level, the structural organization of learning within neural networks themselves remains a crucial determinant in achieving sustainability. Hierarchical decomposition offers a pathway toward energy-efficient AI training, particularly for large-scale models such as foundation models and LLMs, which require massive computational resources \cite{2501.07487v1}.

Traditional deep learning models depend on gradient-based backpropagation, where all model parameters are updated simultaneously across multiple layers. This approach results in high computational costs, excessive memory overhead, and significant energy consumption, making it inefficient for large architectures such as LLMs and transformers. Hierarchical decomposition offers a compelling alternative by partitioning networks into modular learning components, where lower layers focus on fundamental representation learning while higher layers specialize in task-specific adaptation \cite{LLM-in-a-flash}. This design principle is increasingly important in scalable LLM training, where incremental updates and efficient data reuse can significantly reduce computational waste.
Research in multi-scale learning and modular deep networks has demonstrated the benefits of hierarchical architectures. In computer vision, multi-scale learning processes image features at varying resolutions, improving both accuracy and efficiency. For language models, hierarchical structures have been investigated to enhance memory efficiency and modular adaptation, enabling techniques like parameter-efficient fine-tuning (PEFT) to reduce computational overhead in LLM adaptation \cite{2111.05193v2}. Similarly, modular neural networks divide complex architectures into smaller, manageable sub-networks that can be trained independently and later combined for enhanced performance, aligning with recent advances in LLM sparsity-based optimizations \cite{Fire-Flyer_AI-HPC_A_Cost-Effective_Software-Hardware_Co-Design_for_Deep_Learning}.

A primary benefit of hierarchical decomposition is its facilitation of incremental learning, which allows models to adapt to new data without requiring complete retraining. This is particularly useful in large-scale AI applications, where continual adaptation is necessary but complete retraining remains computationally infeasible. Instead of retraining all layers, hierarchical decomposition enables selective updates, preserving previously learned representations while ensuring computational efficiency and adaptability \cite{On_the_Viability_of_Using_LLMs_for_SW_HW_Co-Design_An_Example_in_Designing_CiM_DNN_Accelerators}. For LLMs, this methodology aligns with techniques such as LoRA-based adaptation, where only a subset of model parameters is updated to improve efficiency.
From a hardware perspective, hierarchical decomposition naturally integrates with emerging hardware architectures designed for efficient AI computation. Lower layers, which perform fundamental feature extraction, are well-suited for FPGA acceleration, where parallel processing can be leveraged for low-energy, high-speed computation. Meanwhile, higher layers, which require adaptability, can benefit from dynamic optimization techniques implemented on general-purpose processors. Recent work also highlights how hierarchical memory architectures can improve efficiency in transformer-based models, reducing redundant computations during inference\cite{2501.11006v1}. Additionally, recent research in processing-in-memory (PIM) architectures has shown that reducing memory access overhead can further enhance the efficiency of hierarchical training, as PIM-based accelerators minimize energy-intensive data movement \cite{2412.08490v1}.

Despite these benefits, several challenges remain in making to make hierarchical decomposition a practical and scalable solution. For LLMs, a key challenge is balancing hierarchical updates while preserving long-range dependencies in model representations. Current methods, such as sparse fine-tuning and layer-wise freezing, aim to address this issue, but additional innovations are required to optimize memory efficiency while maintaining model accuracy. Another challenge is determining the optimal division of computational tasks across different layers. While lower layers can leverage direct equation solving for rapid convergence, higher layers must balance adaptability with stability to ensure long-term learning efficiency. Additionally, designing neural network architectures that can seamlessly integrate multiple optimization strategies—such as hardware acceleration for early layers and adaptive tuning for deeper layers—requires advancements in neural architecture search and compiler optimizations \cite{2406.18142v1}.

Another important consideration is preventing fragmentation in the learning process introduced by hierarchical decomposition. Since different parts of the network are optimized separately, there is a risk that information flow between layers may be disrupted. For LLMs, maintaining coherent feature representations across hierarchical layers is crucial, particularly in long-context reasoning tasks. Techniques such as recurrent memory mechanisms and hybrid attention structures are being explored to mitigate this issue \cite{2205.09646v1}. For these concerns, research explored hierarchical feature fusion techniques that allow layers to share information while maintaining computational efficiency \cite{Green_Edge_AI_A_Contemporary_Survey}. By incorporating mechanisms such as attention-based interactions between layers, hierarchical models can maintain strong representation learning capabilities while benefiting from energy-efficient training.

In summary, hierarchical decomposition presents a scalable solution for training deep learning models with reduced energy consumption. By structuring neural networks into distinct levels with specialized optimization strategies, this approach enables efficient training, incremental learning, and better adaptability, particularly in LLMs and large-scale AI models. While existing research has demonstrated its feasibility, further research is required to refine its integration with LLM memory-efficient architectures, edge deployment strategies, and hardware-optimized training pipelines. The next section will introduce our proposed methodology, which builds upon these principles to develop a scalable and energy-efficient AI framework. Moreover, this decomposition paradigm naturally aligns with emerging concepts like Compound LLM frameworks, where two distinct LLM modules are hierarchically organized for specialized optimization.

\section{Compound LLM: An Energy-Efficient Hierarchical Neural Network with Fast FPGA-Based Incremental Learning}

Addressing the scalability and energy efficiency challenges of AI training requires innovative optimization strategies that minimize redundant computations while maintaining model adaptability. We propose a hybrid approach that combines hierarchical decomposition with FPGA-based direct equation solving and incremental learning. This framework is designed to support not only conventional deep learning models but also the expanding class of foundation models and LLMs, which require high computational resources and frequent adaptation.
Our method is motivated by the need for scalable, energy-efficient AI training that overcomes the limitations of traditional optimization techniques. Standard gradient-based training involves iterative weight updates, which lead to excessive power consumption, memory overhead, and computational inefficiencies. While FPGAs have shown promise for accelerating model training with reduced power usage, previous studies have often been limited to smaller models or lacked support for online adaptability. A promising recent method that integrates direct equation solving with FPGA-based incremental learning has shown that eliminating gradient descent can drastically reduce computational costs. However, such approaches encounter challenges when applied to large-scale models like LLMs, due to increasing memory and compute demands. To address these limitations, our framework decomposes the neural network into two functional hierarchies—feature extraction and decision-making— each mapped to hardware-aware optimization strategies. Lower layers leverage FPGA-accelerated equation solving for fast and efficient training, while higher layers are updated incrementally using gradient-based learning, enabling rapid adaptation to new data distributions without full retraining. In addition to our original hierarchical approach, we introduce a Compound LLM framework inspired by advancements in agent-based AI and compound AI architectures. This concept explicitly employs LLM modules at both layers of the hierarchy, offering a more focused and visually interpretable application of our method to LLMs (see Fig.~\ref{fig:compound-llm}). The lower-layer LLM focuses on efficient feature extraction, while the upper-layer LLM handles adaptive decision-making—preserving energy efficiency and scalability.

\subsection{Hierarchical Decomposition Strategy}
Traditional deep learning models update all layers simultaneously using gradient-based optimization, leading to significant computational and memory overhead. To overcome this limitation, we decompose the model into a two-tier architecture, each optimized using different methods suited to their role and resource demands.

\begin{itemize}
\item \textbf{Lower layers (Feature Extraction Layers):} Responsible for capturing reusable, domain-agnostic representations, these layers are optimized through direct equation solving on FPGAs to reduce redundant computations and lowering energy consumption. \item \textbf{Higher layers (Adaptive Decision Layers):} Focused on task-specific adaptation, these layers are updated using incremental learning techniques to efficiently accommodate evolving data distributions.\end{itemize}

This hierarchical strategy enables selective, resource-aware updates—keeping the energy-optimized lower layers static while adaptively tuning only the upper layers.
Such decomposition is particularly effective in modular architectures like LLMs, where separating general-purpose representation learning from task-specific reasoning improves both efficiency and scalability. Moreover, this decomposition principle naturally aligns with emerging concepts like Compound LLM frameworks, where two distinct LLM modules can be hierarchically combined for specialized optimization.

\subsection{FPGA-Based Direct Equation Solving for Efficient Feature Extraction}
\begin{figure}[h]
\centering
\includegraphics[width=0.50\linewidth]{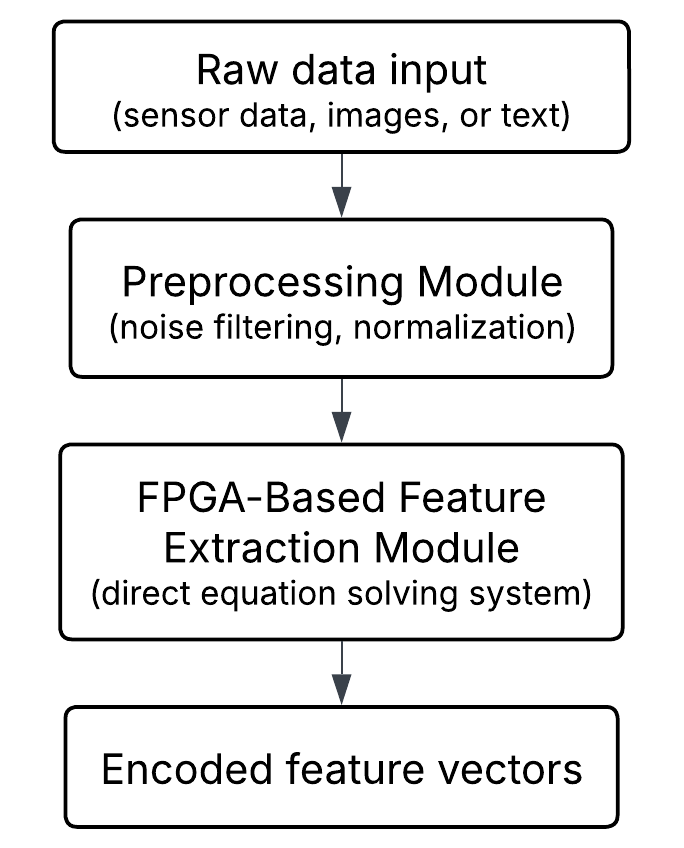}
\caption{Lower-level architecture leveraging FPGA-based direct equation solving for efficient feature extraction.}
\label{fig:lower-layer}
\end{figure}

The lower layers of the proposed model, responsible for fundamental feature extraction, are optimized using direct equation solving on Field-Programmable Gate Arrays (FPGAs). Unlike traditional backpropagation, which updates weights through iterative steps, this method computes optimal parameters in a single pass, significantly reducing training complexity and energy consumption. 
This design corresponds to the energy-efficient component  of our Compound LLM framework, where a compact, energy-efficient LLM agent is assigned to extract reusable representations with minimal cost. An overview of this lower-layer structure is illustrated in Fig.~\ref{fig:lower-layer}.
Given a training dataset $(X, Y)$, the weight matrix is computed as:
\begin{equation}
W = (X^T X + \lambda I)^{-1} X^T Y
\end{equation}
where $\lambda I$ is a small regularization term to enhance numerical stability.

This equation is efficiently solved using FPGA-based matrix inversion and multiplication, leveraging the inherent parallelism of FPGA to accelerate computation while minimizing power consumption.

FPGA-based direct equation solving offers several advantages. First, it significantly enhances energy efficiency by enabling parallel processing with lower power consumption than GPUs-based methods. Second, it eliminates iterative gradient updates, allowing the model to compute optimal parameters in a single step, which leads to faster convergence. Third, this approach is particularly effective for small to mid-sized networks, where direct equation solving remains computationally feasible. When integrated as the lower-layer LLM in our compound framework, this module provides a sustainable foundation for downstream reasoning, while supporting task-level scalability.
While FPGA-based direct equation solving offers significant advantages in energy efficiency and rapid convergence, it faces notable limitations in scalability and practical implementation. As model size grows, solving large systems of equations becomes increasingly resource-intensive, constrained by FPGA on-chip memory and computational complexity, which may limit direct applicability to extremely large models such as GPT-4. Moreover, FPGA development often requires specialized expertise in hardware design and lacks the rapid prototyping flexibility of GPU-based pipelines. These challenges highlight the need for hybrid optimization strategies, such as sparsity-aware FPGA adaptations, and software-hardware co-design frameworks that abstract hardware complexities, thereby improving scalability and accessibility for large-scale AI systems.

As LLMs and foundation models increasingly rely on reusable modules (e.g., encoders or embeddings), these lower-layer components can benefit from energy-efficient FPGA optimization, reducing the cost of fine-tuning and downstream adaptation. However, applying direct equation solving to deep architectures presents challenges. The computational complexity grows exponentially as the depth of the model increases, making it difficult to scale to large neural networks. Additionally, memory constraints limit the feasibility of storing and processing large datasets efficiently. To overcome this, our method decouples the adaptation process, delegating flexibility to higher layers through incremental learning mapped to the second, adaptive LLM agent in our framework.

\subsection{Incremental Learning for Adaptive Model Updates}
\begin{figure}[h]
\centering
\includegraphics[width=0.50\linewidth]{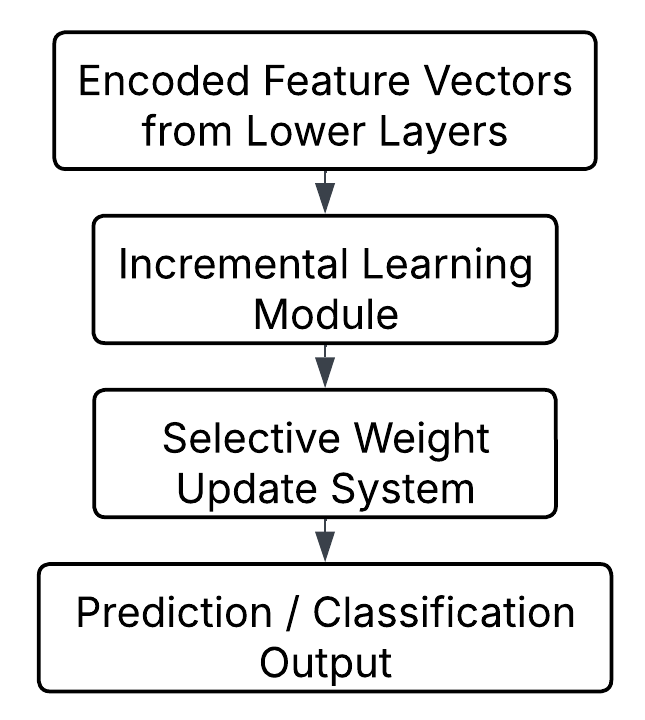}
\caption{Higher-level adaptive learning framework with incremental updates.}
\label{fig:higher-layer}
\end{figure}

The higher layers of deep learning models are responsible for capturing complex representations and must continuously adapt to evolving data distributions. Retraining the entire model for every new batch of data is computationally expensive and inefficient. Instead, we employ incremental learning, allowing selective updates while keeping the lower layers fixed, thereby reducing unnecessary computations and enhancing adaptability. Fig.~\ref{fig:higher-layer} illustrates the structure of the higher-layer incremental learning process.

Given a new batch of data $(X', Y')$, parameter updates for the higher layers are performed as:
\begin{equation}
\theta_H^{(t+1)} = \theta_H^{(t)} + \eta \nabla_{\theta_H} \mathcal{L}(f(X'; \theta), Y')
\end{equation}
where:
\begin{itemize}
\item $\theta_H^{(t)}$ represents parameters at timestep $t$.
\item $\eta$ is the learning rate.
\item $\nabla_{\theta_H} \mathcal{L}$ denotes the gradient of the loss function with respect to the higher-layer parameters.
\end{itemize}
This update strategy enables fine-grained adaptation to new tasks or data streams—an essential feature for deployment in foundation models where continual learning is key. Incremental learning enhances scalability by allowing models to evolve over time without full retraining, significantly reducing computational costs. However, it introduces potential risks such as catastrophic forgetting and model drift during long-term deployment. To mitigate these challenges, stability-preserving strategies such as Elastic Weight Consolidation (EWC) can be incorporated, penalizing changes to parameters critical for previously learned tasks. This regularization ensures a balance between adaptability to new data and retention of prior knowledge, preserving model robustness while maintaining efficiency, particularly in dynamic or resource-constrained environments.

Nevertheless, incremental updates can introduce drift in model weights, leading to catastrophic forgetting. To mitigate this, regularization-based techniques such as Elastic Weight Consolidation (EWC) can be applied, ensuring stability while adapting to new data:
\begin{equation}
\mathcal{L}_{\text{EWC}} = \mathcal{L} + \frac{\lambda}{2} \sum_{i} F_i (\theta_i - \theta_i^*)^2
\end{equation}
where:
\begin{itemize}
\item $\lambda$ controls the trade-off between stability and adaptability.
\item $F_i$ denotes the Fisher Information for parameter $\theta_i$.
\item $\theta_i^*$ is the optimal parameter prior to adaptation.
\end{itemize}
This approach preserves essential knowledge while selectively refining the model based on new inputs. Such stability-preserving mechanisms are crucial when incrementally updating LLMs with domain-specific data.

By combining FPGA-based direct equation solving for lower layers with incremental learning for higher layers, our method strikes a balance between efficiency and adaptability. Lower layers extract fundamental features with minimal computational cost, while higher layers dynamically adjust to new data, ensuring that the model remains effective in changing environments and scalable for large architectures such as foundation models and LLMs.

Our proposed method integrates hierarchical decomposition with FPGA-based direct equation solving and incremental learning to create a sustainable and scalable AI training framework. By structuring the network into two distinct optimization levels, our approach minimizes computational redundancy while maintaining adaptability to evolving data distributions. The lower layers are optimized using FPGA-based direct equation solving, which eliminates the need for iterative backpropagation and significantly lowers energy consumption, making the training process more energy efficient. In contrast, higher layers leverage incremental learning strategies, enabling selective updates that prevent full model retraining and support continual adaptation to new data—a critical requirement for long-term deployment. This dual optimization framework is particularly promising for energy-aware deployment of foundation models and LLMs, where training and inference efficiency are paramount. Additionally, this approach directly aligns with the principles of sustainable AI, addressing both energy efficiency and computational scalability. The combination of FPGA, adaptive learning mechanisms, and hierarchical decomposition ensures that deep learning models can operate with lower resource consumption while maintaining high accuracy. Furthermore, by reducing reliance on traditional GPU-based training pipelines, our method improves hardware-aware AI sustainability, making it particularly advantageous for edge computing and resource-constrained environments. Ultimately, the proposed framework bridges the gap between energy-efficient AI computation and adaptive learning, paving the way for more environmentally responsible and scalable AI solutions.

These two mechanisms—efficient feature extraction via FPGAs and dynamic model adaptation via incremental learning—form the foundation of our Compound LLM framework, which we elaborate on in the following subsection.

\subsection{Compound LLM Architecture}

To further enhance the clarity, practicality, and relevance of our hierarchical framework for large-scale deployment, we introduce a novel Compound LLM structure inspired by recent advances in agent-based AI systems and modular neural architectures. Unlike traditional monolithic LLM structures, the Compound LLM explicitly employs two hierarchically organized LLM modules, each optimized to address different aspects of computational efficiency, adaptability, and sustainability.
The Compound LLM framework consists of two specialized modules, each optimized through distinct hardware-aware and learning strategies:
\begin{itemize} \item \textbf{Lower-layer LLM (Energy-Efficient Feature Extraction Agent)}: This module uses a compact, resource-conscious LLM architecture specifically optimized for foundational representation learning. By employing FPGA-based direct equation solving, it computes optimal parameters in a single step, eliminating the need for backpropagation and significantly reducing both training time and energy consumption. This efficient representation extraction enables fast, low-power operation, particularly advantageous in edge devices or scenarios with strict power constraints. The FPGA implementation also supports parallel computation, enabling real-time inference with low latency.
\item \textbf{Higher-layer LLM (Adaptive Incremental Decision-Making Agent)}: Building upon the distilled representations from the lower layer, this module applies incremental learning techniques to adapt to new and evolving data distributions. Rather than retraining the entire model, this layer performs selective parameter updates using regularization techniques such as Elastic Weight Consolidation (EWC) to preserve prior knowledge. This adaptive mechanism allows the model to respond efficiently to distributional shifts, making it well-suited for dynamic environments where continual learning is required without incurring the cost of full retraining.
\end{itemize}

\begin{figure}[h] \centering \includegraphics[width=0.605\linewidth]{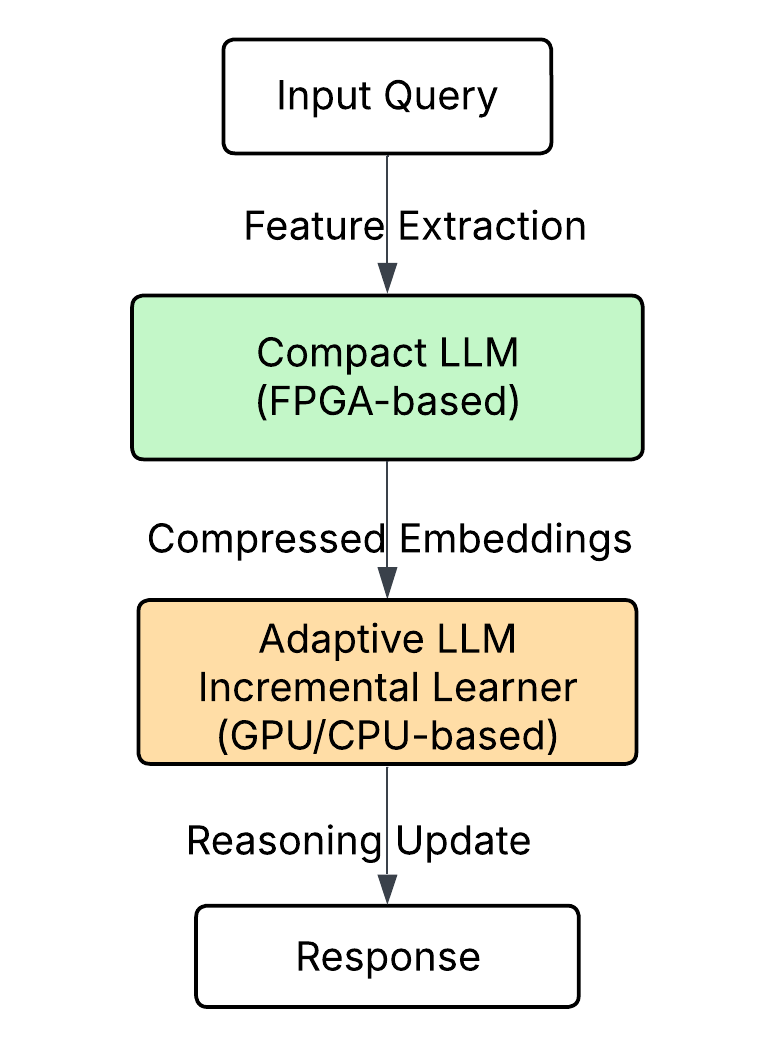} \caption{Hierarchical Compound LLM architecture with hardware-aware optimization and adaptive learning. The lower-level FPGA-accelerated LLM handles efficient representation extraction, while the upper-level LLM conducts task-specific reasoning and incremental updates on conventional processors.} \label{fig:compound-llm} \end{figure}

Fig.~\ref{fig:compound-llm} shows our Compound LLM framework, illustrating the structured interaction between the energy-efficient lower-layer module and the adaptive higher-layer module. The lower layer performs lightweight yet robust feature extraction, which is subsequently passed to the upper layer for task-specific reasoning and real-time model refinement.
The key advantages of our Compound LLM architecture include: 
\begin{itemize} 
\item \textbf{Enhanced Energy Efficiency}: An FPGA-based LLM with offloading early-stage representation learning uses low energy without sacrificing performance.
\item \textbf{Incremental Scalability}: The higher-layer LLM adapts independently through continual learning, eliminating the need for full retraining and enabling modular growth. 
\item \textbf{Optimized Resource Utilization}: The separation of responsibilities between the two modules allows for effective use of heterogeneous hardware, balancing FPGA speed with the flexibility of GPUs/CPUs.
\item \textbf{Robustness to Distributional Shifts}: The incremental learning strategy ensures the model adapts to new patterns while retaining prior knowledge, a crucial property for real-world and streaming-data environments. 
\end{itemize}

\section{Conclusion}
A hierarchical decomposition framework that integrates FPGA-based direct equation solving and incremental learning to enhance energy efficiency and scalability of deep learning models is developed. The new approach with two optimization levels reduces computational overhead, lowers energy consumption, and enables selective model adaptation, addressing the scalability limitations of hardware-accelerated AI.

Our method directly contributes to sustainable AI by reducing power consumption and minimizing reliance on GPU-intensive training pipelines. Through FPGA acceleration and adaptive learning, this framework offers a promising energy-efficient alternative for deep learning applications, particularly in edge computing, low-resource settings, and emerging areas such as LLMs. Additionally, we present the Compound LLM framework, explicitly highlighting LLM-specific optimizations and further demonstrating the practical relevance and scalability of hierarchical sustainable AI.

\section{Future Work}

Future research directions include extending our approach to support real-time LLM inference and fine-tuning, integrating it with hardware-optimized techniques such as neuromorphic computing and Processing-in-Memory (PIM) to further reduce power consumption, and exploring meta-learning strategies to dynamically balance optimization between direct equation solving and incremental learning. Additionally, future work should thoroughly investigate the scalability and practical deployment of the Compound LLM framework, with an emphasis on specialized FPGA accelerations and incremental adaptation strategies for large foundation models and LLMs. As transformer-based models continue to scale in complexity, tailoring hierarchical decomposition strategies to their modular structures could offer new pathways for achieving energy-efficient deployment at scale.

Although our proposed framework offers promising theoretical advancements, its full practical realization requires further empirical validation, scalability optimization for complex LLMs, and real-world deployment studies. These directions are key priorities for future research.

\end{document}